\documentclass[letterpaper, 10 pt, conference]{ieeeconf}
\IEEEoverridecommandlockouts                              
\usepackage{graphics, float, amsfonts, amsmath, amssymb, multirow, verbatim,array, bm}
\usepackage{threeparttable}
\usepackage{algorithm, algpseudocode} 
\usepackage{graphicx}
\usepackage{xcolor}

\title{\LARGE \bf Editing Driver Character: Socially-Controllable Behavior Generation for Interactive Traffic Simulation}
\author{Wei-Jer Chang\textsuperscript{*}, Chen Tang\textsuperscript{*}, Chenran Li, Yeping Hu, Masayoshi Tomizuka, and Wei Zhan
\thanks{\textsuperscript{*} Equal contribution}%
\thanks{W.J. Chang, C. Tang, C. Li, Y. Hu, M. Tomizuka, and W. Zhan are with the Department of Mechanical Engineering, University of California, Berkeley, CA 94720 USA
{\tt\footnotesize \{weijer\_chang, chen\_tang, chenran\_li, yeping\_hu, tomizuka, wzhan\}@berkeley.edu}}%
}

\begin{document}
\maketitle
\thispagestyle{empty}
\pagestyle{empty}

\begin{abstract}
Traffic simulation plays a crucial role in evaluating and improving autonomous driving planning systems. After being deployed on public roads, autonomous vehicles need to interact with human road participants with different \emph{social preferences} (e.g., selfish or courteous human drivers). To ensure that autonomous vehicles take safe and efficient maneuvers in different interactive traffic scenarios, we should be able to evaluate autonomous vehicles against reactive agents with different social characteristics in the simulation environment. We propose a socially-controllable behavior generation (SCBG) model for this purpose, which allows the users to specify the \emph{level of courtesy} of the generated trajectory while ensuring \emph{realistic} and \emph{human-like} trajectory generation through learning from real-world driving data. Specifically, we define a novel and differentiable measure to quantify the level of courtesy of driving behavior, leveraging marginal and conditional behavior prediction models trained from real-world driving data. The proposed courtesy measure allows us to auto-label the courtesy levels of trajectories from real-world driving data and conveniently train an SCBG model generating trajectories based on the input courtesy values. We examined the SCBG model on the Waymo Open Motion Dataset (WOMD) and showed that we were able to control the SCBG model to generate realistic driving behaviors with desired courtesy levels. Interestingly, we found that the SCBG model was able to identify different motion patterns of courteous behaviors according to the scenarios.
\end{abstract}

\section{Introduction}
Simulation plays a critical role in the development and evaluation of autonomous vehicles (AVs). It accelerates the development cycle by enabling efficient evaluation~\cite{9294368} and closed-loop training~\cite{9829243}, which are costly and risky in real-world scenarios. One key component to design for the simulator is the behavior model of the simulated road participants. In conventional driving simulation software, typical methods to synthesize the simulated agents include replaying driving logs~\cite{scheel2022urban} and heuristics-based approaches~\cite{CARLA, SUMO}. Replaying driving logs collected in real-world traffic scenarios ensures \emph{realistic} driving behavior. However, this method does not account for the \emph{reaction} of the simulated agents to the tested AVs. Thus, it is only reliable if the behavior of the AVs stays close to the recorded driving logs, which limits its capability to evaluate AV algorithms in highly interactive scenarios. In contrast, we may leverage domain knowledge to synthesize heuristic-based reactive agents. While the synthesized reactive behavior is sensible, it lacks sufficient realism because these heuristic-based models lack the modeling capacity to simulate sophisticated human-like behavior. 

\begin{figure}[t]
  \centering
  \includegraphics[width=\linewidth]{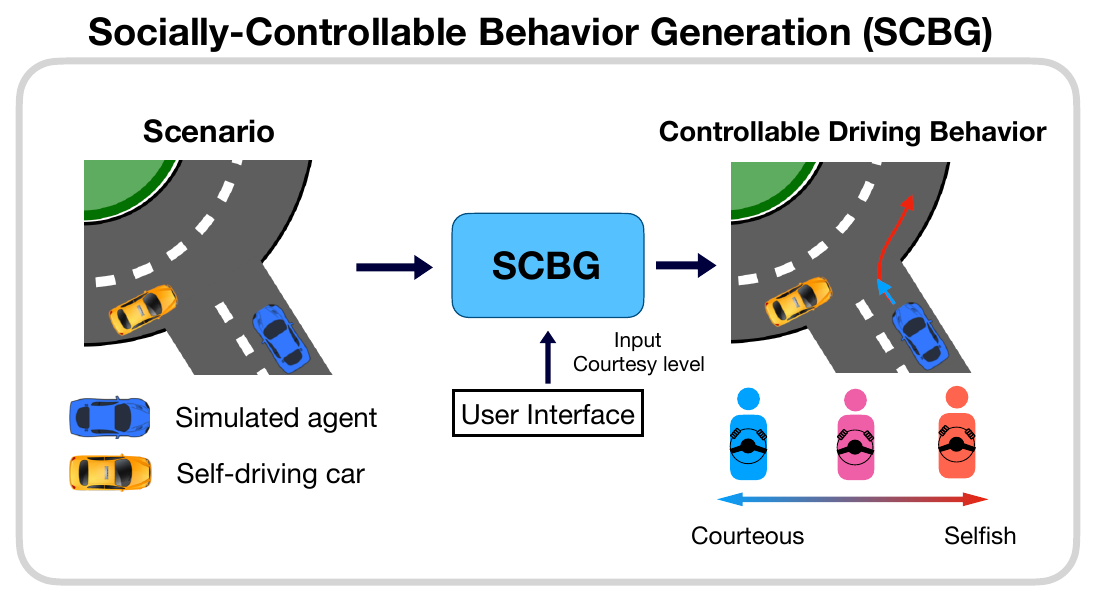}
  \caption{Illustration of the proposed SCBG model. The SCBG model aims to control the driving behavior of a simulated agent based on the input courtesy level. By generating driving behaviors with controlled courtesy levels, the SCBG model could potentially enable more efficient evaluation and training of autonomous driving algorithms in interactive traffic scenarios.}
  \label{fig:illustration}
\end{figure}

To synthesize \emph{realistic} reactive agents, deep learning methods have been proposed to learn reactive driving behavior from large-scale real-world data~\cite{9829243, trafficsim, simnet}. However, such approaches can only synthesize reactive behavior following the training data distribution without a mechanism to \emph{control} the behavior of the reactive agents~\cite{NvidiaGuided}. Recently, several works have attempted to fill in this gap and proposed \emph{controllable} learning-based reactive behavior generation frameworks which allow users to control different aspects of the simulation (e.g., level of safety criticality~\cite{Diverse, STRIVE, advsim}, complying user-specified rules~\cite{NvidiaGuided}). The control mechanisms allow users to specify the simulated scenarios, which could further accelerate the development pipeline. For example, it has been demonstrated that we could evaluate the planner's performance more efficiently by deliberately generating more safety-critical testing scenarios~\cite{Diverse}. 

In this work, we investigate such a controllable learning-based behavior generation framework. Different from prior works, we aim to explicitly control the \emph{social preferences} of the reactive agents. After being deployed on public roads, AVs will encounter human drivers with different driving styles. By specifying the social preferences of the simulated reactive agents, we may have a more comprehensive evaluation of the AVs' capability to handle different human drivers. For instance, an ideal AV should yield to \emph{selfish} drivers to ensure safety but behave less conservatively when encountering \emph{courteous} drivers to ensure efficiency. Also, we may leverage the controllable simulation to design training curricula to accelerate policy training or synthesize driving policies with robust performance for AV as it interacts with human drivers with diverse social characteristics. 

The main challenge lies in \emph{how to quantify and label the social preference of human driving behavior}. Human driving behavior involves sophisticated reasoning procedures which may be characterized from various different aspects. Inspired by prior works~\cite{SVO, courteous_AV}, we choose to control the level of \emph{courtesy} of the reactive agents. Courtesy can be formalized as the change in the expected \emph{reward} of the other agents due to the simulated agent's actions. Intuitively, selfish agents may negatively influence the utilities of other agents, whereas courteous agents take actions that maximize the utilities of other agents. It has been shown as an important factor describing social interaction in driving~\cite{courteous_AV}. Furthermore, we propose a novel data-driven auto-labeling framework that leverages a marginal behavior predictor and a conditional behavior predictor~\cite{M2I, salzmann2020trajectron++, CBP} to estimate the courtesy values of driving behaviors in real-world data. With the auto-labeling framework, we can avoid costly and biased manual efforts in labeling the social characteristics of driving data. 

Given real-world driving data with courtesy value labels, we train a \emph{socially-controllable} behavior generation (SCBG) model synthesizing a simulated agent's future trajectory given an input courtesy value. During training, we leverage the auto-labeling framework to augment the training data with synthesized trajectories. We also utilize the differentiable nature of the auto-labeling operation to design a \emph{courtesy loss}, which directly matches the courtesy values of the generated trajectories with the input courtesy values. Furthermore, we introduce a range predictor to estimate the range of feasible courtesy values in a given scenario, which ensures that the input courtesy value of the SCBG model is feasible during inference time. We examined the proposed method on the Waymo Open Motion Dataset (WOMD)~\cite{ettinger2021large} and showed that we were able to control the SCBG model to generate realistic driving behaviors with desired courtesy levels. Interestingly, we found that the SCBG model was able to identify the different motion patterns of courteous behaviors according to the scenarios. This work is a crucial step toward developing a scalable traffic simulator with socially-controllable and realistic reactive agent models. 

\section{Related Works}
\subsection{Behavior Generation for Traffic Simulation}
The methods for traffic simulations can be broadly categorized into two main groups: heuristic-based and learning-based approaches. In heuristic-based simulations, the reactive agents are controlled with human-specified rules, such as Intelligent Driver Model (IDM)~\cite{CARLA, IDM}. However, these methods lack the modeling capacity required to simulate realistic and human-like behavior. On the other hand, learning-based simulations leverage deep learning-based methods to mimic driving behavior from large-scale driving data~\cite{trafficsim, simnet, igl2022symphony}. For instance, TrafficSim~\cite{trafficsim} trained a variational autoencoder-based model to jointly simulate the agents interacting with each other. Symphony \cite{igl2022symphony} combined parallel beam search and goal conditioning to improve the realism and diversity of the simulated driving behavior.

\subsection{Controllable Behavior Generation}
Recently, several methods have been developed to train controllable and realistic driving behavior generation models from large-scale driving datasets~\cite{STRIVE, advsim, Diverse,NvidiaGuided}. Some of them aim to generate safety-critical scenarios for efficient evaluation. For instance,~\cite{STRIVE} and~\cite{advsim} generate challenging near-collision scenarios using adversarial optimization. \cite{Diverse} proposes a generative model that can produce interactions with varying safety levels controlled by a style coefficient in the latent space. Unlike them, \cite{NvidiaGuided} proposes a controllable behavior generation framework that allows users to specify the desired properties of the trajectories. Our work differs from existing works in that we focus on controlling the \emph{social} characteristics of the generated driving behaviors.

Our work is also related to the literature on modeling social interaction in human driving behavior. For example,~\cite{courteous_AV} formalizes the courtesy level of human drivers as the effect of a driver's behavior on the other drivers' utilities. Based on this,~\cite{letian} further characterizes a human driver's social preferences. They then formulate pairwise interaction as a Stackelberg game where each driver’s utility is a weighted sum of three reward terms for these perspectives. These approaches can be seen as model-based methods for controllable behavior generation. However, they have limited modeling capacity and scalability beyond pairwise interactions. Instead, we aim to explore a data-driven framework that can generate realistic and socially-controllable driving behavior in multi-agent scenarios.

\begin{figure*}[t]
  \centering
  \includegraphics[width=\linewidth]{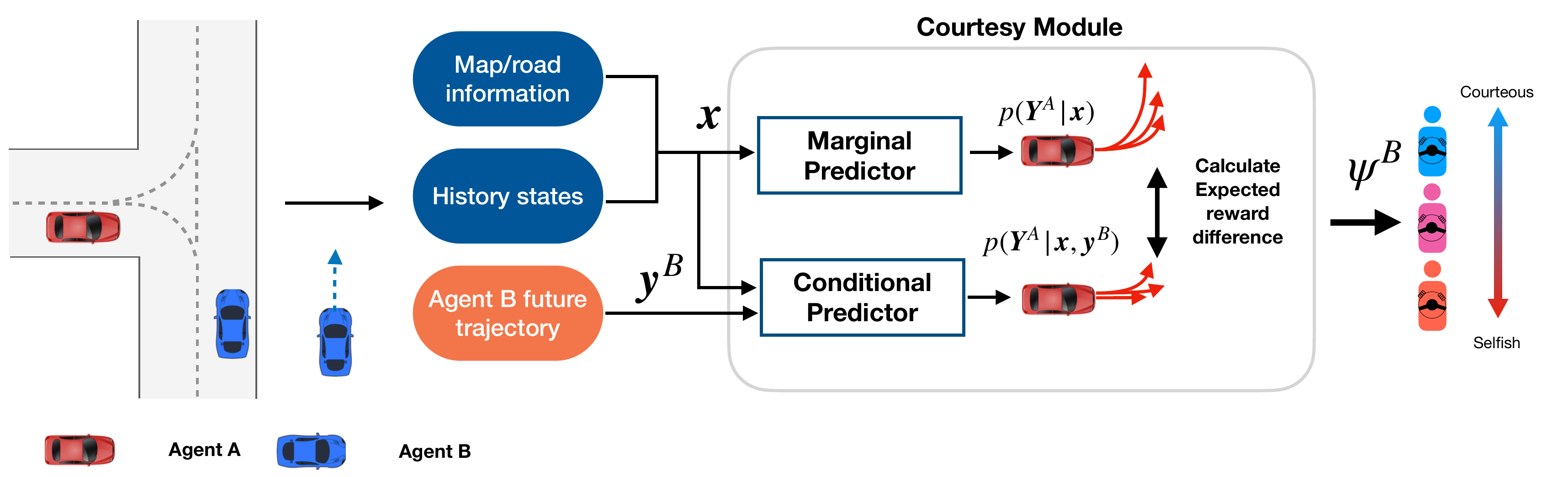}
  \caption{Our proposed auto-labeling framework. The proposed auto-labeling framework estimates and labels the courtesy values of agent B in a sample of interactive pairs by leveraging predictors trained on real-world data. Marginal and conditional trajectory distributions of agent A are estimated by querying the respective predictors, and the difference in the expected reward values under the two distributions is used to quantify the courtesy level. This approach provides an estimate of the courtesy value without requiring humans to manually label the courtesy values.}
  \label{fig:labeling_framework}
\end{figure*}

\section{Socially-Controllable Behavior Generation} \label{sec:method}
In this section, we introduce the Socially-Controllable Behavior Generation (SCBG) framework. In Sec.~\ref{subsec:problem}, we introduce the problem setting considered in this work. In Sec.~\ref{subsec:courtesy_def}, we define a quantitative data-driven measure of courtesy, which is the core element of our SCBG framework. In Sec.~\ref{subsec:labeling}, we introduce the data-driven framework to auto-label the courtesy values of vehicle trajectories from real-world data. In Sec.~\ref{subsec:training}, we explain how we train an SCBG model from real-world data. In Sec.~\ref{subsec:range predictor}, we introduce a courtesy range predictor which is used to control the SCBG model during inference.

\subsection{Problem Formulation}\label{subsec:problem}
We consider a simulated interactive traffic scenario consisting of two vehicles denoted by Vehicle A and Vehicle B, where Vehicle A is controlled by the tested autonomous driving software and Vehicle B is controlled by the behavior generation model we develop. We aim to design a behavior generation model that allows the users to control \emph{how courteous the behavior of Vehicle B is to Vehicle A}. Concretely, we denote the past observation by $\boldsymbol{x}$, which collects the historical trajectories of the two vehicles and other scene information (e.g., map, static or non-interacting objects). Given $\boldsymbol{x}$, the behavior generation model outputs a future trajectory for Vehicle B to follow, denoted by $\boldsymbol{y}^B$. In addition, the model takes a coefficient $\psi$ as input, which controls the level of courtesy of the generated trajectory. Formally, the model is defined as:
\begin{equation}
    \boldsymbol{y}^B = g_{\boldsymbol{\theta}}(\boldsymbol{x}, \psi),\label{eq:generator}
\end{equation}
where $\boldsymbol{\theta}$ denotes the model parameters. We aim to train the model from real-world driving data to ensure the realism of the generated trajectories. However, the courtesy level $\psi$ is a latent variable that is not recorded in the dataset. Thus, we need to define a quantitative measure of courtesy that allows convenient auto-labeling without manual effort. 

\subsection{Quantifying Courtesy}\label{subsec:courtesy_def}
Inspired by~\cite{SVO, courteous_AV}, we formalize courtesy as the change in the expected \emph{reward} of Vehicle A due to the actions of Vehicle B. Formally, given a prospective future trajectory of Vehicle B, $\boldsymbol{y}^B$, we define its level of courtesy as:
\begin{equation}
\psi(\boldsymbol{y}^B) = \mathbb{E}\left[r(\mathbf{Y}^A)\vert \boldsymbol{x},\boldsymbol{y}^B\right] - \mathbb{E}\left[r(\mathbf{Y}^A)\vert \boldsymbol{x}\right], \label{eq:courtesy} 
\end{equation}
where 
\begin{equation}
\begin{aligned}
\mathbb{E}\left[r(\mathbf{Y}^A)\vert \boldsymbol{x},\boldsymbol{y}^B\right] = \int_{\boldsymbol{y}^A} r(\boldsymbol{y}^A)p(\boldsymbol{y}^A|\boldsymbol{x},\boldsymbol{y}^B)d\boldsymbol{y}^A,
\label{eq::Econdition}
\end{aligned}
\end{equation}
\begin{equation}
\begin{aligned}
\mathbb{E}\left[r(\mathbf{Y}^A)\vert \boldsymbol{x}\right] = \int_{\boldsymbol{y}^A} r(\boldsymbol{y}^A)p(\boldsymbol{y}^A|\boldsymbol{x})d\boldsymbol{y}^A. \label{eq::Emarginal}
\end{aligned}
\end{equation}
The variable $\boldsymbol{y}^A$ denotes the future trajectory of Vehicle A. The distribution $p(\boldsymbol{y}^A|\boldsymbol{x}, \boldsymbol{y}^S)$ is the \emph{conditional probability} of Vehicle A's future trajectory, given the observation and Vehicle B's future trajectory. The distribution $p(\boldsymbol{y}^A|\boldsymbol{x})$ is the \emph{marginal probability} of Vehicle A's future trajectory. The reward function $r(\cdot)$ gives the cumulative reward of a given trajectory. In practice, we may specify different reward functions according to the scenarios and the user's needs. 

The term $\mathbb{E}\left[r(\mathbf{Y}^A)\vert \boldsymbol{x}\right]$ indicates the expected reward of Vehicle A under all the possible interactions between the two vehicles, whereas $\mathbb{E}\left[r(\mathbf{Y}^A)\vert \boldsymbol{x},\boldsymbol{y}^B\right]$ indicates the expected reward of Vehicle A if Vehicle B executes a particular trajectory $\boldsymbol{y}^B$. The value $\psi$ then indicates the effect of a given future trajectory of Vehicle B on the expected reward of Vehicle A. A positive $\psi$ indicates a positive effect on the reward and thus implies a courteous and cooperative agent. Conversely, a negative $\psi$ indicates a negative effect on the reward and thus implies a selfish agent.  

\subsection{Auto-Labeling Courtesy}\label{subsec:labeling}
The courtesy value defined above depends on the trajectory distribution of the interacting vehicle. However, we do not have access to the ground-truth trajectory distribution of the vehicles appearing in real-world data. Instead, we propose to estimate and auto-label the courtesy values leveraging trajectory predictors trained from real-world data. The auto-labeling framework is illustrated in Fig.~\ref{fig:labeling_framework}. Given a sample from the dataset, i.e., $\boldsymbol{x}, \boldsymbol{y}^B\sim \mathcal{D}$, we query a marginal predictor for an estimated marginal distribution $p(\boldsymbol{y}^A|\boldsymbol{x})$, and query a conditional predictor for an estimated conditional distribution $p(\boldsymbol{y}^A|\boldsymbol{x},\boldsymbol{y^B})$. With the estimated distributions, we can then estimate the courtesy value of $\boldsymbol{y}^B$ using Eqn.~\eqref{eq:courtesy}. We denote the overall courtesy computation operation by $J_{\boldsymbol{\phi}}(\cdot,\cdot)$, where $\boldsymbol{\phi}$ denotes the parameters of the predictors:
\begin{equation}
    \psi = J_{\boldsymbol{\phi}}(\boldsymbol{x}, \boldsymbol{y}^B).
\end{equation}

Note that the auto-labeling procedure does not impose any restrictions on the prediction models used. In our experiment, we adopted the state-of-the-art Multipath++~\cite{multipath++} as the backbone prediction model, where the predicted trajectory distribution is represented as a Gaussian Mixture Model (GMM). The original Multipath++ model is designed for marginal prediction. We accommodate it to support both marginal and conditional predictions following the practice in~\cite{CBP}. Specifically, we add an additional future encoder to encode $\boldsymbol{y}^B$. Under the conditional prediction mode, the encoded embedding is fused with the embedding of the other input features. Under the marginal prediction mode, we turn off the future encoder so that the model does not take $\boldsymbol{y}^B$ as input. We follow the training scheme in~\cite{CBP} to train the model for both inference modes simultaneously. 

\begin{figure}[t]
  \centering
  \includegraphics[width=\linewidth]{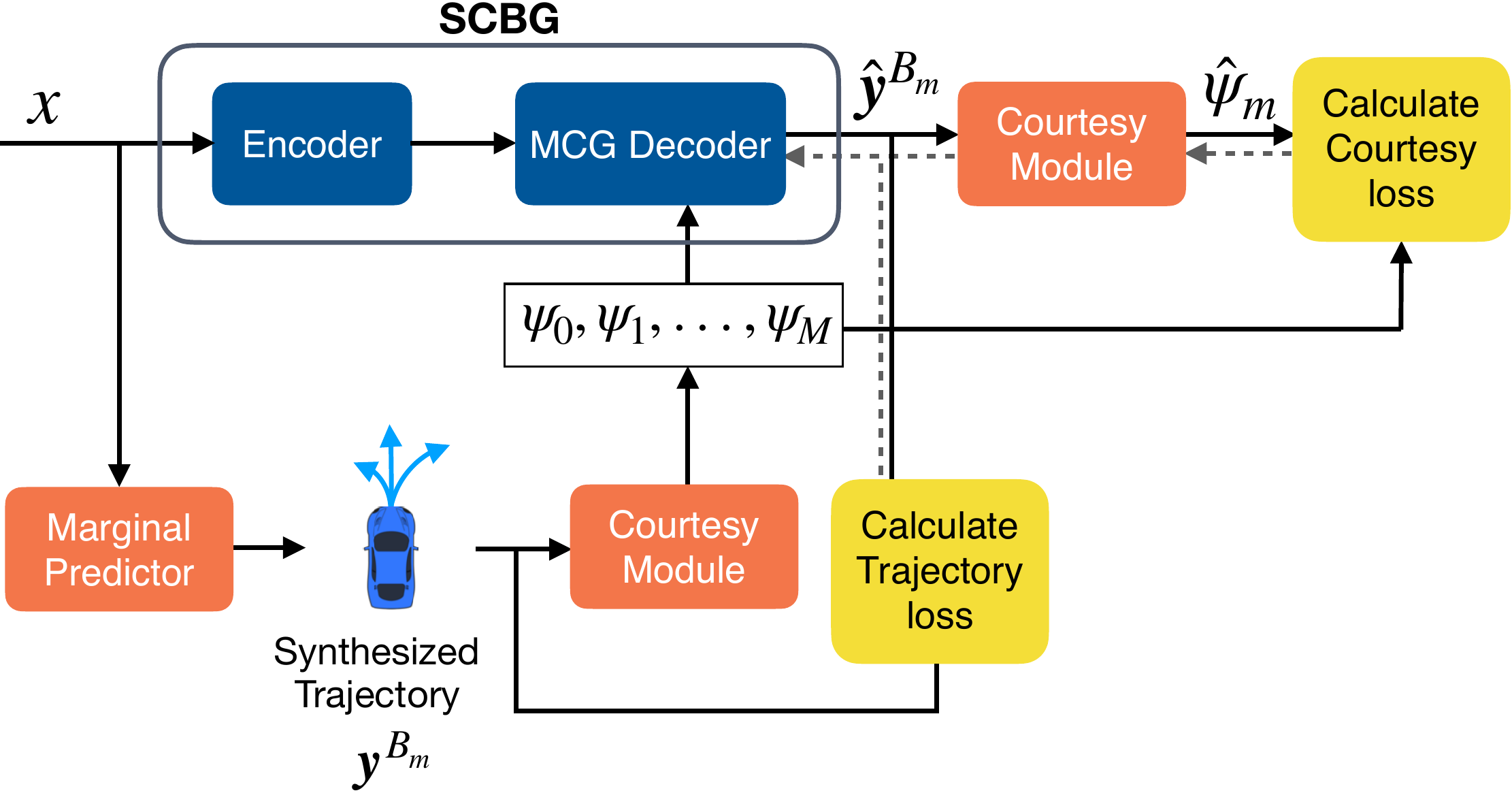}
  \caption{The SCBG model architecture and training pipeline. The SCBG model takes as input the scene observation $\boldsymbol{x}$ and conditions on the courtesy level to produce a trajectory $\boldsymbol{\hat{y}^{B_m}}$. The model is trained by matching the generated trajectories and their courtesy values with the labeled ones. To enhance the model performance, we leverage a marginal predictor to augment the training data with predicted trajectories.}
  \label{fig:training_pipline}
\end{figure}

\subsection{Socially-Controllable Behavior Generation Model}\label{subsec:training}
We now present the SCBG model architecture and its training pipeline, which are summarized in Fig.~\ref{fig:training_pipline}. We implement the SCBG model bases on a Multipath++ backbone, which consists of a scene encoder and a multi-context gating (MCG) trajectory decoder. The original MCG decoder in Multipath++ utilizes a set of learned anchor embeddings for multimodal trajectory prediction. Since our SCBG model is designed for closed-loop simulation tasks, it is crucially important to ensure the long-term stability of the closed-loop behavior~\cite{trafficsim, chang2022analyzing, bits}. Modeling the multimodality at the trajectory level could be troublesome for closed-loop simulation because we need to ensure the consistency in modality across nearby timesteps, which requires non-trivial adaptation of the MCG decoder. For simplification, we remove the anchor embeddings and let the model output a single-modal trajectory given an input courtesy value in this work. In the future, we plan to introduce a high-level goal inference module, such as the one in~\cite{bits}, to model the multimodality when extending the current framework for socially-controllable closed-loop simulation. In particular, we want to highlight two key components of the training process.  


\subsubsection{Data Augmentation} Real-world data can only provide one sample of the future trajectory per scenario, resulting in one sampled behavior that corresponds to a single courtesy value. However, to achieve controllable behavior generation, the model needs to be reliable under a diverse set of input courtesy values. To address this challenge, we augment the training dataset with synthesized trajectories sampled from the marginal predictor. While in auto-labeling, we query the marginal predictor for $p(\boldsymbol{y}^A\vert \boldsymbol{x})$, in this context, we query the marginal predictor for $p(\boldsymbol{y}^B\vert \boldsymbol{x})$ and sample $M$ trajectories from the predicted distribution. By sampling from the marginal distribution, we augment the dataset with plausible trajectories of diverse courtesy levels. Our experiments demonstrate that data augmentation plays a critical role in ensuring the model generates trajectories that correspond to the input courtesy value.

\subsubsection{Loss Function} During training, we use a loss function that consists of two parts. The first part computes the error between the generated and labeled trajectories. Specifically, given a sample from the dataset, $\boldsymbol{x},\boldsymbol{y}^{B,0}\sim \mathcal{D}$, we denote the $M$ synthesized trajectories as $\boldsymbol{y}^{B,m}, m=1,\dots,M$. And we denote the labeled courtesy values for $\boldsymbol{y}^{B,m}$ as $\psi_m$. The trajectory loss, $L_{traj}$, is then defined as:
\begin{equation}
L_{traj} = l\left(\boldsymbol{y}^{B,0},{\boldsymbol{\hat{y}}^{B,0}}\right) + \alpha{\sum_{m=1}^M l\left(\boldsymbol{y}^{B,m},{\boldsymbol{\hat{y}}^{B,m}}\right)},
\end{equation}
where $\boldsymbol{\hat{y}}^{B,m}$ is the trajectory generated by the SCBG model given the courtesy value $\psi_m$, and $l$ is a differentiable loss function, such as mean squared error. The coefficient $\alpha$ is a hyperparameter used to balance the losses between the ground truth and synthesized trajectories. And we use Huber loss as the loss function $l(\cdot,\cdot)$. 

The second part of the loss function compares the input courtesy values against the courtesy values of the generated trajectories, which ensures the generated trajectories indeed match the input courtesy values. The courtesy loss $L_{courtesy}$ is defined as:
\begin{equation}
    L_{court}=\sum_{m=0}^M \|\psi_m - J_{\boldsymbol{\phi}}(\boldsymbol{x},\boldsymbol{y}^{B,m})\|^2.
\end{equation}
Since the courtesy computation module $J_{\boldsymbol{\phi}}(\cdot,\cdot)$ is differentiable, we can directly leverage it to compute the courtesy loss and backpropagate through it for gradient computation. The overall loss function is then defined as:
\begin{equation}
    L= L_{traj}+\beta{L_{court}},
\end{equation}
where the coefficient $\beta$ is a hyperparameter balancing the trajectory loss and the courtesy loss. 
\begin{figure}[t]\
  \centering
  \includegraphics[width=\linewidth]{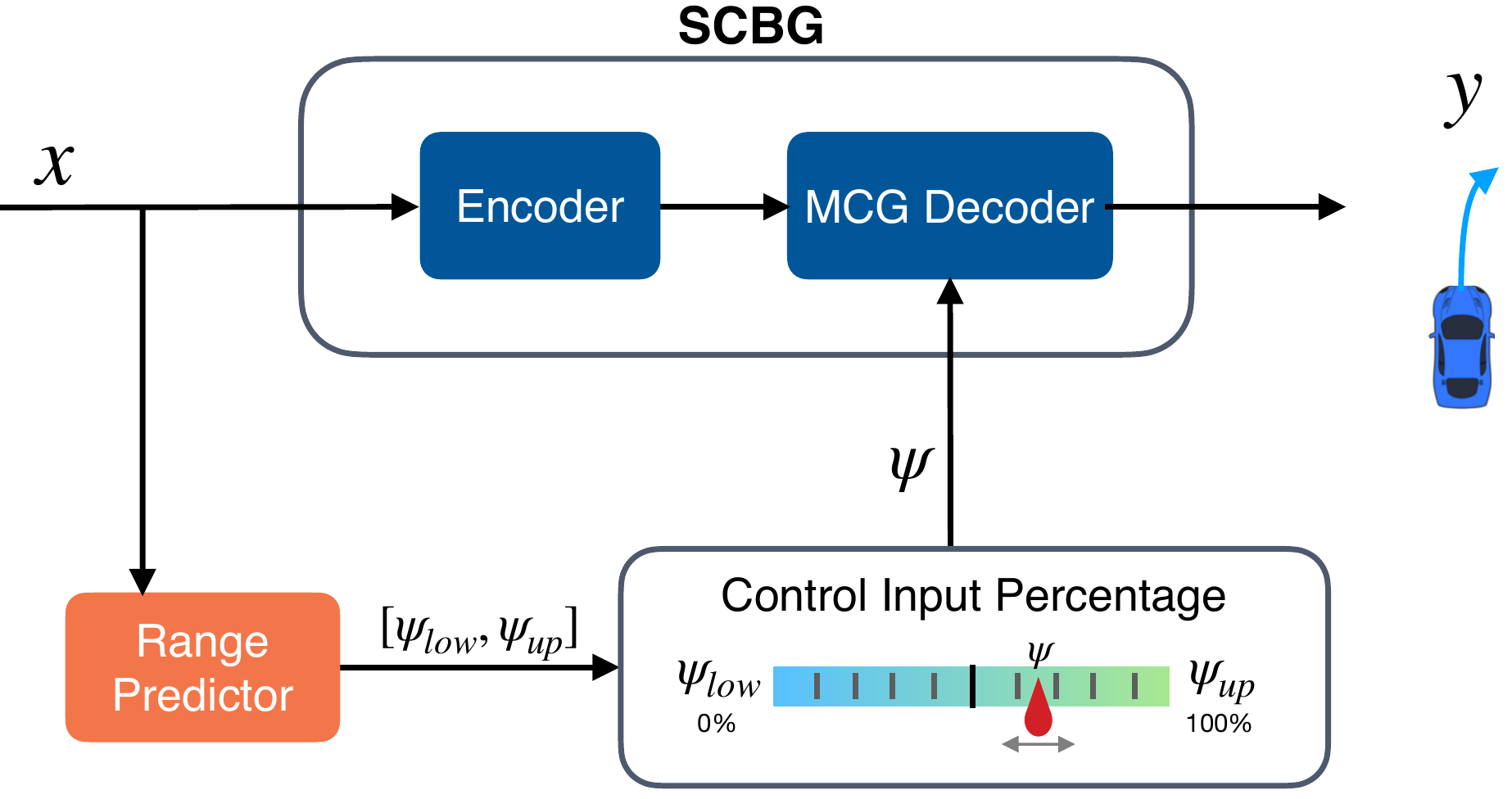}
  \caption{The paradigm of behavior generation at inference time. The range predictor predicts the feasible range of courtesy values for a given scenario. This range is then used to scale the user input percentage to the appropriate input value, which is fed into the SCBG model.}
  \label{fig:range_predictor}
\end{figure}

\subsection{Courtesy Range Predictor}\label{subsec:range predictor}
Since the reward distribution varies across scenarios, so does the range of feasible courtesy values. As a result, feeding arbitrary courtesy values to the SCBG model may lead to out-of-distribution input which results in unreliable generated behavior. To this end, we train a courtesy range predictor to predict the interval of feasible courtesy values for a given scenario. Since we do not have access to the ground-truth distributions, we use quantile regression~\cite{quantile_regression} to estimate the statistics of the courtesy value from data. As shown in Fig.~\ref{fig:range_predictor}, the estimated range is used to normalize the courtesy values across scenarios so that the user only needs to specify the level of courtesy as a value from the unit interval. The range predictor shares the encoder of the SCBG model. During training, we freeze the parameters of the encoder and train a new decoder to predict the 0.1 and 0.9 quantiles of $\psi$. We follow the practice in~\cite{quantile_NN} and use the pinball loss function to train the range predictor: 
\begin{equation}
    L_{\tau} = \begin{cases} (\tau - 1) \cdot (\psi - \hat{\psi}_\tau), & \text{if } \psi < \hat{\psi_\tau} \\ \tau \cdot ( \psi -\hat{\psi_\tau}) & \text{otherwise}, \end{cases}
\end{equation}
where $\tau$ denotes the target quantile (i.e., 0.1 and 0.9) and $\hat{\psi}_\tau$ denotes the predicted courtesy value at the $\tau$-quantile. 


\section{Experiments}
In this section, we conduct experiments to validate our proposed SCBG framework on real-world driving data. 

\subsection{Dataset}
We use the Waymo Open Motion Dataset (WOMD)~\cite{ettinger2021large}. The dataset provides a subset with labels identifying a pair of interacting agents in the scenarios. We refer to this subset as the interactive subset. In our work, we focus on vehicle-to-vehicle interaction and vehicle behavior generation while we still include the historical trajectories of the other types of agents (i.e., pedestrians and cyclists) in the model input. 

\subsection{Training Marginal and Conditional Predictors}\label{exp:predictor}
We build our prediction model on the open-source implementation of Multipath++~\cite{multipath_implementation}. We follow the practice in~\cite{CBP} and train a prediction model supporting both marginal and conditional predictions during inference. We report the prediction errors (measurement time of 8s) on the interactive subset of the validation set in Table~\ref{table:predictor_performance}. The evaluation metrics are defined as in~\cite{waymo_dataset} and computed with respect to six predicted trajectory samples, except for the $\mathrm{ADE}$ metric. The $\mathrm{ADE}$ metric measures the error between the ground-truth trajectory and the predicted sample with the highest probability. Since the SCBG model generates a single trajectory, we list the $\mathrm{ADE}$ values here as references to validate the realism of the trajectories generated by the SCBG model, which will be discussed later in Sec.~\ref{exp:CBG_evaluation}. Notably, the prediction errors are comparable with the results reported in~\cite{multipath_implementation}, which achieves the $3^{\mathrm{rd}}$ place in Waymo Motion Prediction Challenge 2022. 


\subsection{Auto-labeling Courtesy Value}\label{exp:distribution_of_court}
\begin{figure}[t]
  \centering
  \includegraphics[width=0.8\linewidth]{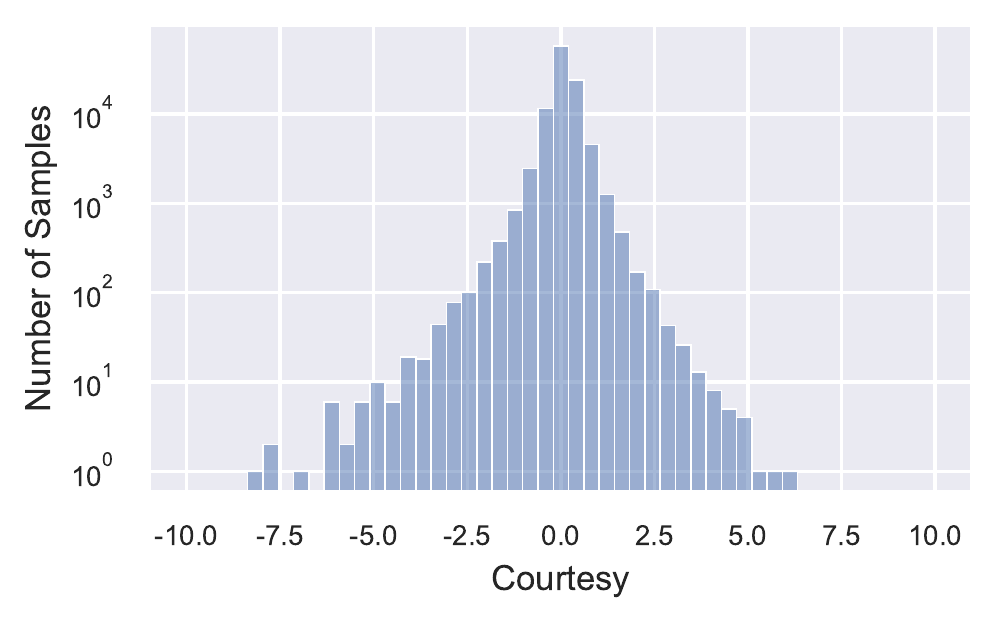}
  \caption{Histogram of courtesy values on the interactive subset of WOMD.}
  \label{fig:histogram}
\end{figure}

Using the trained predictor, we auto-labeled the courtesy values of the data, following the method discussed in Sec.~\ref{subsec:courtesy_def}. Since the courtesy values are non-trivial only in interactive scenarios, we auto-labeled the courtesy values and trained the SCBG model on the interactive subset. In our experiment, we use average speed as the reward function when quantifying the level of courtesy in Eqn.~\eqref{eq:courtesy}. The average speed indicates the agent's progress along its route, which is the primary driving target for on-road driving. Fig.~\ref{fig:histogram} shows the histogram of the extracted courtesy values on the interactive subset. It is worth noting that the histogram concentrates at zero. One reason is that there exist many non-interactive cases even in the interactive subset. Besides, the logged trajectory in the dataset does not necessarily affect the expected utility of the interacting agent. 


\begin{table}[t]
\centering
\caption{Prediction Model Performance}\label{table:predictor_performance}
\begin{tabular}{|c|c|c|c|c|c|} \hline
Mode                                                     & \begin{tabular}[c]{@{}c@{}}ADE\\(m)\end{tabular} & \begin{tabular}[c]{@{}c@{}}minADE \\(m)\end{tabular} & \begin{tabular}[c]{@{}c@{}}minFDE \\(m)\end{tabular} & \begin{tabular}[c]{@{}c@{}}MR \\(\%)\end{tabular} & mAP    \\ \hline
\begin{tabular}[c]{@{}c@{}}Marginal \\ \end{tabular}    & 3.18                                             & 1.16                                                 & 2.52                                                 & 20.6                                              & 0.262  \\ \hline
\begin{tabular}[c]{@{}c@{}}Conditional \\ \end{tabular} & 3.13                                             & 1.11                                                 & 2.35                                                 & 19.5                                              & 0.274  \\ \hline
\end{tabular}
\end{table}

\subsection{Socially-Controllable Behavior Generation}\label{exp:CBG_evaluation}
\subsubsection{Implementation Details} Since a large portion of the data has relatively low absolute courtesy values (Fig.~\ref{fig:histogram}), we split the data into two subsets with a threshold absolute courtesy value of 2. During training, we sample 50\% of the batch data from each subset to avoid overwhelming the training data with trivial samples. To train the SCBG model, we load the encoder parameters from the Multipath++ model and only train the decoder for SCBG.

\subsubsection{Evaluation Metrics} We want to validate that the proposed SCBG model can achieve \emph{socially-controllable} trajectory generation and that the generated trajectories are \emph{realistic}. To evaluate the model’s controllability, we compare the courtesy values of the generated trajectories with the input courtesy values. Since we do not have access to the ground-truth trajectory distributions, we leverage the auto-labeling method to define a data-driven evaluation metric for controllability, denoted by $\mathrm{CourtesyMSE}$, which is the mean squared error (MSE) between the input and auto-labeled courtesy values of the generated trajectory: 
\begin{equation}
    \mathrm{CourtesyMSE} = \frac{1}{N}\sum_{n=1}^N \frac{1}{|\Psi_n|}\sum_{\psi_{n,i}\in {\Psi_n}}(\psi_{n,i} - \hat{\psi}_{n,i})^2,
\end{equation}
where $\hat{\psi}_{n,i}=J_{\boldsymbol{\phi}}\left(\boldsymbol{x}_n,g_{\theta}(\boldsymbol{x}_n, \psi_{n,i})\right)$ and $N$ is the number of samples in the dataset. $\Psi_n$ is a set of selected courtesy values of interest. We consider three strategies to define $\Psi_n$ for different perspectives of evaluation:
\begin{itemize}
    \item \emph{Data}: We define the set $\Psi_n$ with the courtesy values of the ground-truth and augmented trajectories:
    \begin{equation*}
        \Psi_{n,data} = \left\{J_{\phi}\left(\boldsymbol{x}_n,\boldsymbol{y}^{B,m}_n\right)\right\}_{m=0}^M.
    \end{equation*}
    The resulting metric quantifies controllability over input courtesy values following the data distribution, eliminating the influence of infeasible input courtesy values. 

    \item \emph{Range}: We use the range predictor to predict the 0.1 and 0.9 quantiles of the courtesy values and interpolate between the predicted quantiles with a fixed interval:
    \begin{equation*}
        \Psi_{n,range} = \left\{\hat{\psi}_{0.1}(\boldsymbol{x}_n), \hat{\psi}_{0.1+\delta\tau}(\boldsymbol{x}_n),\cdots, \hat{\psi}_{0.9}(\boldsymbol{x}_n)\right\}.
    \end{equation*}
    The resulting metric reflects the controllability of the SCBG model in a practical setting where the range of feasible courtesy values is unknown. 

    \item \emph{Arbitrary}: The input courtesy values are selected by interpolating between the minimum and maximum courtesy values of the entire dataset, denoted by $\psi_{\mathrm{min}}$ and $\psi_{\mathrm{max}}$, in a sample-agnostic way:
    \begin{equation*}
        \Psi_{arbitrary}=\{\psi_{\mathrm{min}},\psi_{\mathrm{min}}+\delta{\psi},\cdots,\psi_{\mathrm{max}}\}.
    \end{equation*}
    The resulting metric serves as a baseline showing the model performance without the range predictor. 
\end{itemize}

To evaluate realism, we follow the common practice~\cite{simnet, NvidiaGuided} to compare the generated trajectory against the ground truth from data. Specifically, we compare the trajectory generated with the courtesy value of the ground-truth trajectory against the ground-truth trajectory. We use $\mathrm{ADE}$ to quantify the trajectory error and define the metric as follows:
\begin{equation}
    \mathrm{TrajADE} = \frac{1}{TN}\sum_{n=1}^N\sum_{t=1}^{T} || \boldsymbol{y}^{B,0}_{n,t} - \boldsymbol{\hat{y}}_t^{B,0}||_2.
\end{equation}
We can then evaluate the realism of the generated trajectories by comparing $\mathrm{TrajADE}$ with $\mathrm{ADE}$ reported in Table~\ref{table:predictor_performance}.

\begin{figure}[t]\
  \centering
  \includegraphics[width=0.9\linewidth]{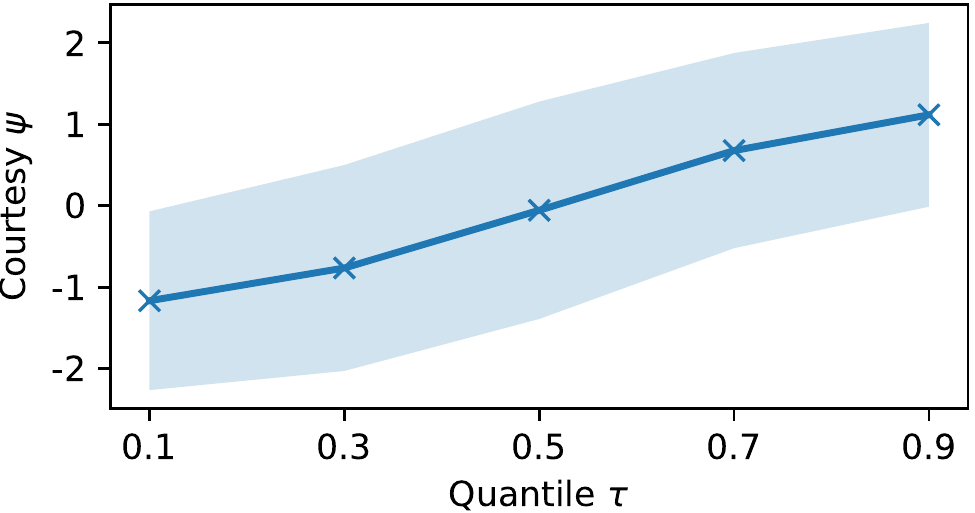}
  \caption{The relationship between input quantile and the courtesy value of the generated trajectory in the high-courtesy subset. The courtesy value increases as the input quantile increases, demonstrating the controllability of our SCBG model.}
  \label{fig:quantile_to_courtesy}
\end{figure}

\subsection{Quantitative Results}
Table~\ref{table:scbg_result} summarizes the results of three model variants: 1) the baseline variant without data augmentation or courtesy loss; 2) the baseline model with data augmentation; and 3) the complete SCBG model with both data augmentation and courtesy loss. The results validate the effectiveness of the proposed data augmentation scheme and the courtesy loss. Both modules significantly reduce courtesy and trajectory errors, implying improved controllability and realism. Interestingly, the courtesy loss helps reduce $\mathrm{TrajADE}$ even though it does not directly penalize trajectory errors, which shows that matching the courtesy values provides informative supervision signals that guide the model to better capture the driving behaviors during training.

Table~\ref{table:scbg_result} also highlights the important role of the range predictor by comparing the courtesy errors under different strategies for constructing $\Psi_n$. Without the range predictor, the courtesy error significantly increases (i.e., when $\Psi_{n}=\Psi_{arbitrary}$) because of the out-of-distribution input courtesy values. Meanwhile, the courtesy error with the range predictor (i.e., when $\Psi_{n}=\Psi_{n, range}$) is comparable to the courtesy error when the courtesy values from the data are used (i.e., when $\Psi_{n}=\Psi_{n, data}$).

Overall, the proposed SCGB model enables socially controllable and realistic driving behavior generation, which is validated by the low $\mathrm{CourtesyMSE}$ value of $0.120$ and a $\mathrm{TrajADE}$ value smaller than the $\mathrm{ADE}$s of the trajectory prediction models reported in Table~\ref{table:predictor_performance}. We further illustrate the model's controllability in the practical setting where users give quantile commands as in Fig.~\ref{fig:range_predictor}. As shown in Fig.~\ref{fig:quantile_to_courtesy}, the courtesy value of the generated trajectory effectively increases with the quantile input. The correlation coefficient between these two values is $0.60$ on the entire interactive validation subset and $0.79$ on the high-courtesy subset (i.e., data with absolute courtesy values larger than 2). 

\begin{table*}[hbtp]
\centering
\caption{Socially-Controllable Behavior Generation Evaluation Results}\label{table:scbg_result}
\centering
\begin{threeparttable}
\begin{tabular}{|c|c|c|c|c|} \hline
\multirow{2}{*}{model} & \multicolumn{3}{c|}{$\mathrm{CourtesyMSE}$} & \multirow{2}{*}{$\mathrm{TrajADE}$}\\\cline{2-4}
& \multicolumn{1}{c}{\emph{data}} & \multicolumn{1}{c}{\emph{range}} & \emph{arbitrary} & \\ \hline
\textcircled{1} & $0.062 \pm 0.321$ & $0.154 \pm 0.470$ & $10.35 \pm 9.40$ & $3.27 \pm 2.60$\\ \hline
\textcircled{1}+\textcircled{2} & $0.045 \pm 0.252$ & $0.137 \pm 0.467$ & $9.97 \pm 9.26$ & $3.22 \pm 2.91$\\ \hline
\multicolumn{1}{|c|}{\textcircled{1}+\textcircled{2}+\textcircled{3}} & \multicolumn{1}{l|}{$0.034\pm 0.172$} & \multicolumn{1}{l|}{$0.120\pm 0.376$} & $9.77\pm 9.09$ & $3.02\pm 2.72$\\ \hline
\end{tabular}
\begin{tablenotes}
\item[]\textcircled{1}: Baseline Model \qquad \textcircled{2}: Data Augmentation \qquad \textcircled{3}: Courtesy Loss
\end{tablenotes}
\end{threeparttable}
\end{table*}

\begin{figure*}[t]
  \centering
  \includegraphics[width=\linewidth]{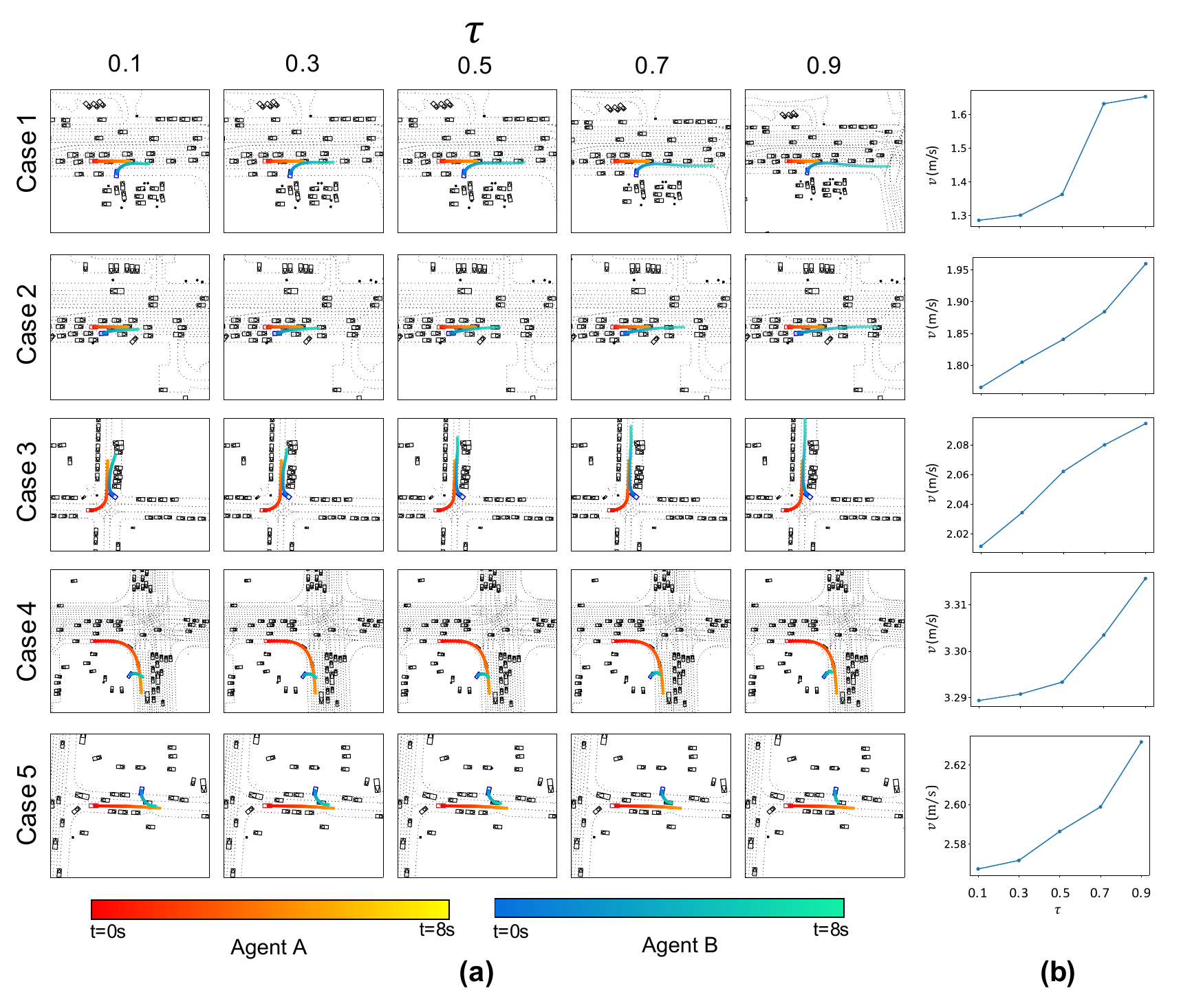}
  \caption{(a) Visualization of the generated trajectories by control courtesy value. The blue line indicates the trajectory of the controlled agent B, while the red line represents the ground truth trajectory of agent A. This visualization shows that as the input courtesy value increases, agent B increases agent A's reward by either changing lanes (Case 1), increasing its speed (Cases 2-3), or yielding (Cases 4-5). (b) Relationship between the input courtesy quantiles and the predicted average speed of agent A, given agent B's trajectory.}\label{fig:visualization}
\end{figure*}

\subsection{Qualitative Analysis}
In this section, we visualize some representative examples showing that the SCBG is able to identify and generate different courteous behaviors according to the scenarios. In~\ref{fig:visualization}, we plot the trajectories generated for the controlled agent B with different input courtesy quantiles. We also plot Agent A's ground-truth trajectories observed in the dataset to visualize the nominal behavior of Agent A. For example, the controlled agent attempts to merge into the lane where Agent A is driving in Cases 1-3. The controlled agent accelerates to prevent blocking Agent A when the courtesy level increases. Interestingly, when the input quantiles are 0.7 and 0.9 in Case 1, the controlled agent further switches its lane to the right, creating more space for Agent A. In contrast, the controlled agent slows down and yields when the courtesy level increases in Cases 4-5. Despite the diverse behavior patterns observed in the visualization, all types of courteous behavior generated by the SCBG model result in a higher reward for the other agent (Fig~\ref{fig:visualization}(b)). It demonstrates the ability of the SCBG model to generate controllable courteous behaviors according to the social context. 

\section{Discussion and Limitation}
We demonstrate that the proposed SCBG framework can control the courtesy level of the generated driving behavior, which is a crucial first step toward achieving socially controllable traffic simulation. The proposed SCBG framework offers a key advantage in that it can be easily scaled to handle multi-agent interactions by calculating the influence of the controlled agent on all the other agents. In future work, we will extend SCBG to a socially-controllable closed-loop simulation framework that enables simulated agents with specified social preferences to interact with the test AVs. One key step is to condition the SCBG model on a goal or reference path to enhance long-horizon closed-loop stability~\cite{igl2022symphony, bits}. Afterward, we are interested in exploring how to leverage a socially-controllable traffic simulator to accelerate policy training and robustify the trained policy. One limitation of the current framework is that the auto-labeling method is affected by the causality issue of conditional behavior prediction~\cite{CBP, tang2022interventional}. In particular, a conditional prediction model cannot differentiate between the correlation and causation of two agents' trajectories. One potential solution is incorporating prior knowledge of causal relations when designing the closed-loop simulation~\cite{M2I}. 

\section{Conclusion}
In this study, we introduce socially-controllable behavior generation (SCBG), a model capable of generating realistic driving behavior corresponding to a desired level of courtesy. The proposed method is empowered by a novel data-driven quantification of courtesy in socially interactive traffic scenarios, which allows us to auto-label the latent courtesy values of real-world driving data. We present a novel training algorithm to train the SCBG model from large-scale real-world driving data. In particular, we introduce a data augmentation scheme and a novel courtesy loss to improve the controllability and realism of the trained model. We showed that we were able to control the SCBG model
to generate realistic driving behaviors with desired courtesy levels. In future work, we will extend SCBG to a socially-controllable closed-loop simulation framework and explore its application in closed-loop policy training and evaluation. 

\section*{ACKNOWLEDGMENT}
The authors would like to thank Prof. Anca Dragan, Lingfeng Sun, and Hengbo Ma for their insightful suggestions. This work was supported by Hong Kong Centre for Logistics Robotics.

\bibliography{IROS2023.bib}
\bibliographystyle{IEEEtran}

\end{document}